# Improving Low-resource Reading Comprehension via Cross-lingual Transposition Rethinking

Gaochen Wu[1], Bin Xu[1], Yuxin Qin[2], Fei Kong[3], Bangchang Liu[3],

Hongwen Zhao[3], Dejie Chang[3]

[1,2]Computer Science and Technology, Tsinghua University, Beijing, China
[3]Beijing MoreHealth Technology Group Co. Ltd
[1]{wgc2019, xubin}@tsinghua.edu.cn, [2]{tyx16}@mail.tsinghua.edu.cn
[3]{kongfei, liubangchang, zhaohongwen, changdejie}@miao.cn

## Abstract

Extractive Reading Comprehension (ERC) has made tremendous advances enabled by the availability of large-scale high-quality ERC training data. Despite of such rapid progress and widespread application, the datasets in languages other than high-resource languages such as English remain scarce. To address this issue, we propose a **C**ross-**L**ingual **T**ransposition Re**T**hinking (XLTT) model by modelling existing high-quality extractive reading comprehension datasets in a multilingual environment. To be specific, we present multilingual adaptive attention (MAA) to combine intra-attention and inter-attention to learn more general generalizable semantic and lexical knowledge from each pair of language families. Furthermore, to make full use of existing datasets, we adopt a new training framework to train our model by calculating task-level similarities between each existing dataset and target dataset. The experimental results show that our XLTT model surpasses six baselines on two multilingual ERC benchmarks, especially more effective for low-resource languages with 3.9 and 4.1 average improvement in F1 and EM, respectively.

## 1 Introduction

Reading Comprehension (RC) is a comprehensive practically-useful task which also can be used to evaluate how well computer systems understand human languages. Due to a large number of high-quality span-extraction RC datasets released in recent years, such as SQuAD v1.1 (Rajpurkar et al., 2016), HotpotQA (Yang et al., 2018), CMRC 2018 (Cui et al., 2019b), DRCD (Shao et al., 2018), and so on, span-extractive RC task has become an enormously popular research topic in NLP and a large number of neural-based RC models have pushed the boundaries of numerous extractive RC datasets. However, these studies have been essentially monolingual and largely focused on English.

Though tremendous advances in popular and realistic span-extraction QA have been made by the NLP community in recent years, there are still two main challenges making training extractive QA systems more difficult: (1) Large-scale high-quality datasets in other languages remain scarce, even for relatively high-resource languages such as Chinese. (2) Collecting such a sufficient amount of training data for all the world's languages is costly and even impossible. Furthermore, we argue that while large-scale supervised training datasets (e.g., WikiQA (Yang et al., 2015), ImageNet (Deng et al., 2009)) are extremely useful in the past decade, but they are too expensive and not strictly necessary in the future. An alternative approach to constructing large scale high-quality monolingual datasets is to develop multilingual modeling systems which can generalize to a wide range of the world's languages whether training datasets are available or not in target languages.

In this year or so, we have seen a few research works on cross-lingual extractive RC modeling. Asai et al. (2018) proposed the first zero-shot extractive RC system for Japanese and French with no extractive training datasets available, instead by an English extractive RC model (e.g., BiDAF (Seo et al., 2017) and improved BiDAF (Clark et al., 2017; Peters et al., 2018) and an attention-based neural machine translation model. Cui et al. (2019a) develop several back-translation approaches and Dual BERT (Devlin et al., 2018) for cross-lingual machine reading comprehension task. However, (1) firstly construct multilingual parallel corpora by translating large existing high-quality extractive

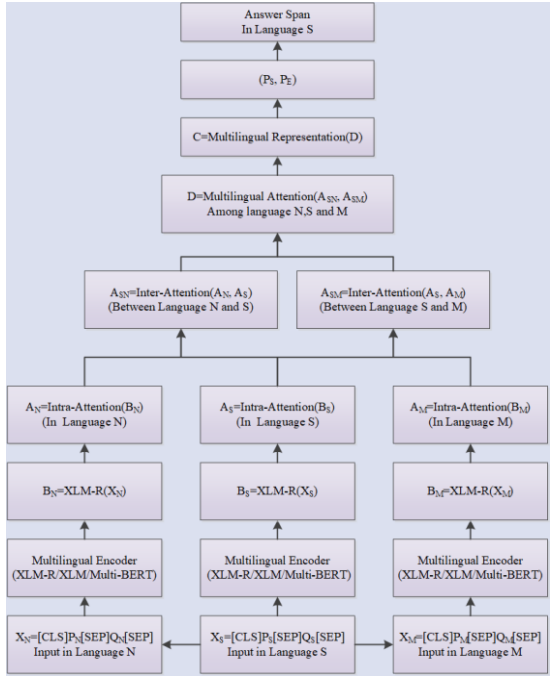

(a) The overall architecture of XLTT using multilingual adaptive attention mechanism with intra-attention, inter-attention and multilingual attention mechanism.

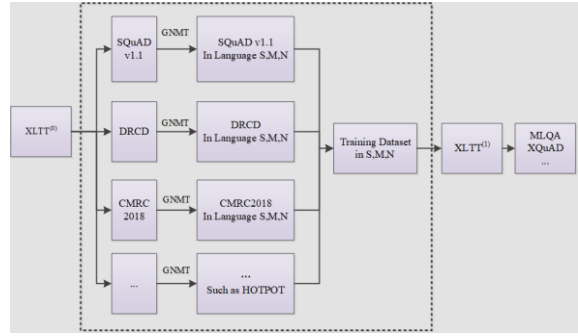

(b) The training framework of XLTT using multiple existing datasets such as SQuAD v1.1, DRCD, CMRC2018

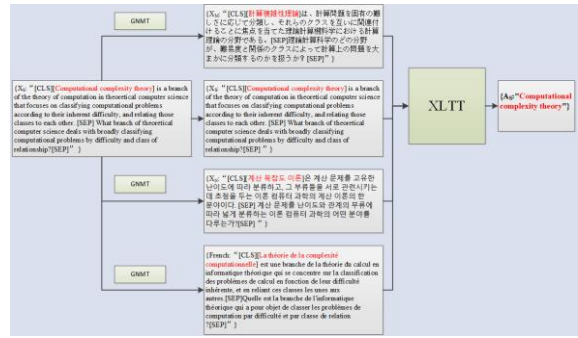

(c) An input example constructed with GNMT from SQuAD v1.1 of XLTT shown in (a)

Figure 1: The overall architecture and training framework of XLTT.

Dual BERT considers training data in a bilingual context to learn semantic relations. (2) Moreover, Dual BERT is proposed only for the conditions when there are training data available in target languages. Although interesting attempts have been made, multilingual extractive RC task still has not been well-addressed. Furthermore, complex linguistic phenomena and typological diversities (Comrie et al., 2015) make developing general multilingual extractive RC models much more challenging.

In order to speedup multilingual extractive RC modeling research, there are several datasets and benchmarks constructed in 2020 or so, including XQuAD (Artetxe et al., 2019), MLQA (Lewis et al., 2019), TYDIQA (Clark et al., 2020), XTREME (Hu et al., 2020), and so on, their state-of-the-art generalization performance is far behind training-language results, showing huge room to improve.

In this paper, we propose a multilingual span-extractive RC modeling approach named **XLTT**, denoting cross-lingual transposition rethinking. The overall architecture and training framework of XLTT is shown as in Figure 1. Specifically, we

RC datasets from rich-resource languages like English into different language families such as Japanese (Kaiser et al., 2013)) and Korean (Han et al., 2005) with SOTA neural machine translation (e.g., GNMT (Wu et al., 2016)). It should be noted that although GNMT is much more powerful, we introduce auxiliary loss for considering the quality of the translated data. Secondly, we adopt multilingual adaptive attention (MAA) mechanism to combine intra-attention and inter-attention which can enable our extractive RC model to learn more general abstract semantic and lexical knowledge from parallel instances in each pair of representative language (e.g., Korean) and pivot language (e.g., English), which is beneficial to generalize to low-resource languages.

To investigate the effectiveness of our proposed approach, we carry out extensive experiments on MLQA and XQuAD, two multilingual extractive reading comprehension datasets without providing training data. The experimental results show that our XLTT model achieves better performance than six baseline models, especially effective for low-resource languages with 5.1 and 4.5 average improvement in F1 and EM on MLQA, 2.7 and 3.6 average improvement in F1 and EM on XQuAD, respectively.

To sum up, we make the following contributions: **(1)** We propose a cross-lingual transposition rethinking (XLTT) approach by leveraging the existing large-scale high-quality ERC datasets in a multilingual context for solving low-resource ERC. **(2)** We propose multilingual adaptive attention (MAA) mechanism by combining intra-attention and inter-attention which can enable our extractive RC model to capture more general generalizable semantic and lexical information from each pair of language families, which is beneficial to generalize to low-resource languages. We also put forward a new training framework to train our ERC model by computing the task-level similarities between each existing training dataset and target dataset.

**(3)** Experimental results show that our proposed model outperforms performance of 6 baselines on 2 multilingual extractive reading comprehension benchmarks, which demonstrate the effectiveness of our multilingual modeling approach.

## 2 Related Work

Reading Comprehension (RC) has been a popular and genuinely-useful NLP research domain in recent years. A large number of datasets are available to study RC from various perspectives, for example, including span-extraction reading comprehension (Wang et al., 2017; Seo et al., 2017; Chen et al., 2017; Wang et al., 2017; Yu et al., 2018), multi-hop reading comprehension (Ding et al., 2019; Zhao et al., 2020; Asai et al., 2020), multi-hop complex KBQA (Zhang et al., 2018; Zhou et al., 2018), open-domain QA (Yang et al., 2019; Lee et al., 2018), dialogue QA (Guo et al., 2018; See et al., 2017), and many more. Due to the availability of large-scale high-quality RC datasets, the NLP community has enjoyed huge progress. Benefiting from created large-scale potentially practically-useful extractive RC datasets such as SQuAD v1.1 (Rajpurkar et al., 2016), TriviaQA (Joshi et al., 2017), NewsQA (Trischler et al., 2016), SQuAD 2.0 (Rajpurkar et al., 2018) and NQ (Kwiatkowski et al., 2019), along with powerful pretraining language models including BERT (Devlin et al., 2018), XLNet (Yang et al., 2019), GPT (Radford et al., 2018), RoBERTa (Liu et al., 2019), and best NLP technologies like attention mechanism (Bahdanau et al., 2014), Transformer (Vaswani et al., 2017), GCN (Kipf et al., 2016), extractive RC has become a dominant paradigm and massive progress has been achieved by the community in the past several years. Despite such popularity and widespread application, **(1)** extractive RC datasets in languages other than English are comparatively rare. Moreover, **(2)** collecting such large-scale high-quality datasets is difficult and costly. Furthermore, **(3)** these studies have been monolingual and largely focused on English. Extractive RC in other languages has not been well-solved.

### 2.1 Multilingual Extractive RC Datasets

In order to accelerate research progress in the field of multilingual extractive RC, several potentially practically-useful extractive RC datasets have been released in this year or so. Lewis et al. (2019) present MLQA, a multi-way parallel extractive question answering evaluation benchmark in seven languages. Clark et al. (2020) propose TYDI QA, a QA dataset covering eleven typologically diverse languages with 204K QA pairs. Artetxe et al. (2019) create XQuAD, a dataset of 1190 SQuAD v1.1 QA pairs from 240 paragraphs by translating the instances into ten languages. Hu, J. et al (2020) release XTREME, a massively multilingual multi-task benchmark to verify and validate cross-lingual generalization capability of multilingual models. The goals of these proposed multilingual datasets and benchmarks are to encourage and speedup research on developing high-quality multilingual question answering systems to generalize across a large number of the world's languages with a wide variety of complex linguistic phenomena and to further accelerate research progress on models that generalize well across the linguistic phenomena and data scenarios of the world's languages. Due to huge challenges arising from typological diversity, generalization capabilities of multilingual models are far behind training-language performance and still have significant room for improvement.

### 2.2 Multilingual Extractive RC Modeling

The increasing interest in cross-lingual extractive RC modeling has been seen in this year or so. We observe a few recently-published cross-lingual extractive RC research works briefly introduced below. Asai et al. (2018) develop the first zero-shot extractive RC system for non-English languages (e.g., Japanese and French), instead by an English extractive RC model and an attention-based neural machine translation (NMT) model. They evaluated their proposed zero-shot extractive RC system on small Japanese and French datasets translated from the S QuAD v1.1 development set. However, the generalization results of their proposed RC system are quite low. In addition, their proposed attention-based answer alignment method between various language families with an astonishing breadth of complex diverse linguistic phenomena are different and could result in additional noise.

Cui et al. (2019) propose several zero-shot back-translation models and Dual BERT for cross-lingual machine reading comprehension. However, back-translation approaches have to align the translated answer with the source answer span in the original paragraph to meet the requirement of extractive reading comprehension. Dual BERT is proposed for improving Chinese extractive reading comprehension task with a certain amount of training data. Moreover, Dual BERT only models the <Passage, Question, Answer> in a bilingual environment to learn the hidden semantic relations and is not a general multilingual ERC model.

In summary, multilingual ERC modeling still has not been well-studied. To address the scarce availability of ERC training data in low-resource languages and overcome challenges from a large variety of complex linguistic phenomena and scenarios, we propose XLTT shown in Figure 1, a cross-lingual transposition rethinking approach. The experimental results show that our approach obtains better performance than six baselines on two multilingual ERC benchmarks.

## 3 Problem Definition

### 3.1 Cross-Lingual Generalization (XLG)

In this paper, we focus on addressing low-resource multilingual extractive reading comprehension like MLQA without training data in target languages, in this work low-resource languages. We propose a cross-lingual generalization (XLG) task, which requires training a model with existing training data ($P_S$, $Q_S$, $A_S$) in high-resource languages other than target languages like Hindi. Development data in high-resource language S is used for picking hyper-parameters. At test time, the trained model is required to extract answer $A_T$ given context $P_T$ and question $Q_T$ in low-resource language T. It should be noted that we only use the supervised signals in high-resource language to train the model. We provide an example shown in (c) of Figure 1.

## 4 Model

In this section, we turn to detail our proposed XLTT illustrated in Figure 1, mainly consisting of the following 3 steps:

(1) For existing extractive reading comprehension datasets, we firstly construct multilingual parallel corpora in multiple language families such as (English, Japanese, Korean, …) using GNMT.

(2) Secondly, we utilize multilingual adaptive attention to learn linguistics-related semantic and lexical knowledge in a multilingual environment for generalizing to low-resource languages.

(3) Then, we calculate task-level cosine similarities between each training dataset and target dataset for using multiple training datasets.

### 4.1 Building Multilingual Parallel Corpora

In this study, we choose different language families from the 11 typologically diverse languages covered by the TYDI QA benchmark to evaluate our proposed XLTT modeling approach. At the same time, to guarantee the zero-shot evaluation, in which representative languages we choose for training multilingual models are also different from target test languages contained by MLQA and XQuAD. Therefore, during the training stage, we choose (English, Japanese, Korean) as language combination. In the test stage, we select (target language such as Hindi, Japanese, Korean).

For the SQuAD v1.1 dataset, we translate to Japanese and Korean. For the CMRC2018 and DRCD, we translate to (English, Japanese, Korean) and discard the Chinese instances due to MLQA and XQuAD including Chinese.

Formally, given an extractive QA instance (e.g., <Passage, Question, Answer>) in high-resource language S (e.g., English) represented as <$P_S$, $Q_S$, $A_S$>, we organize each source language question-passage pair as an input token sequence $X_S$ for multilingual encoder (e.g., multilingual BERT and XLM-R), as follows:

$$X_S = [CLS]Q_S[SEP]P_S[SEP] \qquad (1)$$

Similarly, we can obtain constructed multilingual parallel question-passage pairs of $X_S$ in different language families represented as input token sequences, such as $X_M$ in the language M and $X_N$ in the language N as follows:

$$X_M = [CLS]Q_M[SEP]P_M[SEP] \qquad (2)$$

$$X_N = [CLS]Q_N[SEP]P_N[SEP] \quad (3)$$

## 4.2 Multilingual Adaptive Attention Reader

In this part, we firstly use a shared multilingual encoder such as Multi-BERT and XLM-R, to encode the constructed multilingual parallel corpus to obtain contextualized vector representations. Then we simultaneously model training datasets in a multilingual environment by using multilingual adaptive attention (MAA) mechanism.

**Multilingual Encoding:** To be specific, we use the input token sequence X (e.g., $X_S$, $X_M$ and $X_N$) for multilingual encoder to obtain contextualized representations B (e.g., $B_S$, $B_M$ and $B_N$, respectively) as follows:

$$B = XLMR(X) \quad (4)$$
$$B_S \in R^{L_S*h}, B_M \in R^{L_M*h}, B_N \in R^{L_N*h}$$

Where L indicates the length of the input token sequence and h represents the dimensionality of the hidden state (e.g., 768 for multilingual BERT, 1152 for XLM-R).

**Multilingual Intra/Inter Attention:** Typically, intra-attention mechanism is used to relate different positions of a single sequence in order to calculate more accurate representation of the sequence. Intra-attention has already been used successfully in a variety of NLP tasks including extractive RC. In our multilingual RC modeling approach, to enrich the language representation for final prediction, we firstly take advantage of intra-adaptive attention to filter the irrelevant part of each representation (e.g., $B_S$, $B_M$ and $B_N$) in order to make the attention more precise and then apply the softmax function to obtain self-attended representations as in Eqn. (5), (6) and (7):

$$A_S = \text{softmax}(B_S \cdot B_S^T), A_S \in R^{L_S*h} \quad (5)$$

$$A_M = \text{softmax}(B_M \cdot B_M^T), A_M \in R^{L_M*h} \quad (6)$$

$$A_N = \text{softmax}(B_N \cdot B_N^T), A_N \in R^{L_N*h} \quad (7)$$

Then, we use multi-head inter-attention to further enhance the final representation by modeling the linguistics-related semantic relations between the passage and question in each pair of languages (i.e., the pivot language S and one of various language families, such as language M or N). we calculate inter-attention of each pair of languages (i.e., S and M, S and N) as in Eqn. (8) and (9):

$$A_{SM} = B_S \cdot B_M^T, A_{SM} \in R^{L_S*L_M} \quad (8)$$

$$A_{SN} = B_S \cdot B_N^T, A_{SN} \in R^{L_S*L_N} \quad (9)$$

To utilize the benefit of both intra-attention and inter-attention, we take advantage of multilingual adaptive attention mechanism and calculate the dot product between intra-adaptive attention and all inter-attention representations to obtain attended representations in the following 4 equations:

$$\tilde{A}_{SM} = A_S \cdot A_{SM} \cdot A_M^T, \tilde{A}_{SM} \in R^{L_S*L_M} \quad (10)$$

$$\tilde{A}_{SN} = A_S \cdot A_{SN} \cdot A_N^T, \tilde{A}_{SN} \in R^{L_S*L_N} \quad (11)$$

$$C'_M = \text{softmax}(\tilde{A}_{SM}) \cdot B_M, C'_M \in R^{L_S*h} \quad (12)$$

$$C'_N = \text{softmax}(\tilde{A}_{SN}) \cdot B_N, C'_N \in R^{L_S*h} \quad (13)$$

Furthermore, we propose multilingual attention in order to capture more general semantic knowledge from various languages families by concatenating all attended representations in above Eqn. (12) and (13). Multilingual attention allows the model to jointly attend to useful information from different representation subspaces of different languages.

$$Multilingual(S,M,N,...) = Concat(C'_M, C'_N,...) \quad (14)$$

In this work, we only consider pivot language S and two language families M and N to form a 3-language multilingual context, so the multilingual attention representation can be simplified as in Eqn. (15):

$$C' = Multilingual(S,M,N)$$
$$= Concat(C'_M, C'_N), \quad (15)$$
$$C' \in R^{L_S*2h}$$

After getting multilingual attention representation, we use another fully connected layer along with residual connection and normalization layer to get final knowledge-enhanced source representation as in Eqn. (17). Finally, we use $G_S$ as the multilingual adaptive attention representation for final answer prediction.

$$C = W_C C' + b_C, W_C \in R^{2h*h}, C \in R^{L_S*h} \quad (16)$$

$$G_S = concat[B_S, LayerNorm(B_S + C)], \quad (17)$$
$$G_S \in R^{L_S*2h}$$

### 4.3 Task-level Similarity

For each training dataset $D_i$, we first collect all its questions to form a large question document $Q_i$, and then use the TF-IDF to represent the $D_i$ dataset and target dataset $D_t$ such as MLQA (using the English test set). Finally, we directly use the task-level similarity $S_{ti}$ between $D_i$ and $D_t$ as the cosine similarity as the weight of the dataset $D_i$ during the training stage based on their TF-IDF in Eqn (18):

$$\text{Weight}_{ti} = \text{cosine}(\text{TFIDF}(Q_i), \text{TFIDF}(Q_t)) \quad (18)$$

Then, we normalize the above calculated similarity using Eqn (19).

$$\text{Weight}_{ti} = \text{softmax}(\text{Weight}_{ti}) \quad (19)$$

Note that we only use the question words to calculate the similarity.

**Loss Function**: We use Eqn. (17) to $G_{Si}$ and use $G_{Si}$ to calculate standard cross entropy loss for the answer span in the source training language in Eqn. (20), (21) and (22).

$$P_{Si}^s = \text{softmax}(W_S^s G_{Si} + b_S^s) \quad (20)$$

$$P_{Si}^e = \text{softmax}(W_S^e G_{Si} + b_S^e) \quad (21)$$

$$L_S = -\sum_{i \in S} \text{Weight}_{ti} \cdot \left( \frac{1}{|Si|} \sum_{k=1}^{K} \left( y_{Si}^s \log(P_{Si}^s) + y_{Si}^e \log(P_{Si}^e) \right) \right) \quad (22)$$

Where $S_i$ denotes the i-th training dataset in language $S$ such as SQuAD v1.1.

In order to estimate how the translated instances assemble the original instances, we take advantage of cosine semantic similarity $\alpha_R$ obtained in Eqn. (23) between the golden span representations in source $S$ and representative $R$ languages denoted as $H_S$ and $H_R$ described in (Cui et al., 2018), we are not going to detail how to calculate $H_S$ and $H_R$ in this paper.

$$\alpha_R = \max\{0, \text{Cos} < H_S, H_R >\} \quad (23)$$

$$L_R^{GNMT} = -\frac{1}{K} \sum_{k=1}^{K} \left( y_R^s \log(P_R^s) + y_R^e \log(P_R^e) \right) \quad (24)$$

$$L_{Objective} = L_S + \sum_R \alpha_R L_R^{GNMT} \quad (25)$$

We also calculate the auxiliary cross-entropy loss in Eqn. (24) for translated samples in $R$, where a $\alpha_R$ parameter is applied to this loss. The overall objective function is calculated in Eqn. (25).

## 5 Experimental Details

We now turn toward how to evaluate our proposed multilingual modeling approach using MLQA and XQuAD with the zero-shot evaluation.

| MLQA | Development | Test | Total |
|---|---|---|---|
| English (en) | 1148 | 11590 | 12738 |
| German (de) | 512 | 4517 | 5029 |
| Spanish (es) | 500 | 5253 | 5753 |
| Arabic (ar) | 517 | 5335 | 5852 |
| Chinese (zh) | 504 | 5137 | 5641 |
| Vietnamese(vi) | 511 | 5495 | 6006 |
| Hindi (hi) | 507 | 4918 | 5425 |

Table 1: Number of instances in MLQA

| SQuAD v1.1 | Training | Development | Test |
|---|---|---|---|
| Num.Instances | 87,599 | 10,577 | 9,533 |

Table 2: Statistics of SQuAD v1.1

| Dataset | Training | Development | Test |
|---|---|---|---|
| CMRC2018 | 10,321 | 3,219 | 4,895 |
| DRCD | 26,936 | 3,524 | 3,493 |

Table 3: Statistics of CMRC2018 and DRCD

### 5.1 Implementation Details

We use ADAM with weight decay optimizer using an initial learning rate of 4e-5 and adopt cosine learning rate decay scheme instead of the original linear decay, which we found it beneficial for stabilizing results. All systems are trained on multiple hosts and multiple graphic units of GPU Cloud platform with Tesla V100 with 32 GB RAM. In this paper, we choose SQuAD v1.1, CMRC2018 and DRCD as our original training data to train our XLTT model, and use the English development data for picking hyper-parameters and evaluate the model on the MLQA and XQuAD test set.

### 5.2 Datasets

**XQuAD** (Artetxe et al., 2019) is a benchmark dataset for evaluating cross-lingual span-extractive QA performance. The dataset consists of a subset of 240 paragraphs and 1190 question-answer pairs from the development set of SQuAD v1.1 together with professional translations into 10 languages.

| F1/EM | English | Spanish | German | Arabic | Hindi | Vietnamese | Mandarin | Avg |
|---|---|---|---|---|---|---|---|---|
| BERT-Large (Lewis et al., 2019) | 80.2/67.4 | | | | | | | |
| Multi-BERT (Hu, et al., 2020) | 80.2/67.0 | 67.4/49.2 | 59.0/43.8 | 52.3/34.6 | 50.2/35.3 | 61.2/40.7 | 59.6/38.6 | 61.4/44.2 |
| XLM (Lewis et al., 2019) | 74.9/62.4 | 68.0/49.8 | 62.2/47.6 | 54.8/36.3 | 48.8/27.3 | 61.4/41.8 | 61.1/39.6 | 61.6/43.5 |
| XLM-R (Hu, et al., 2020) | 83.5 / 70.6 | 74.1/56.6 | 70.1/54.9 | 66.6/47.1 | 70.6/53.1 | 74.0/52.9 | 67.1/44.4 | 72.3/54.2 |
| Translate-Test, BERT-Large (Lewis et al., 2019) | | 65.4/44.0 | 57.9/41.8 | 33.6/20.4 | 23.8/18.9 | 58.2/33.2 | 44.2/20.3 | 47.2/29.8 |
| Translate-Train, Multi-BERT (Lewis et al., 2019) | | 53.9/37.4 | 62.0/47.5 | 51.8/33.2 | 55.0/40.0 | 62.0/43.1 | 61.4/39.5 | 57.7/40.1 |
| Translate-Train, XLM (Hu, et al., 2020) | | 65.2/47.8 | 61.4/46.7 | 54.0/34.4 | 50.7/33.4 | 59.3/39.4 | 59.8/37.9 | 58.4/39.9 |
| XLTT with Multi-BERT (ours) | 81.0/68.1 | 67.1/49.2 | 60.0/46.2 | 54.8/36.7 | 50.7/35.7 | 63.9/44.2 | 65.1/44.0 | 63.2/46.3 |
| XLTT with XLM-R (ours) | 83.9/71.2 | 76.7/59.4 | 72.2/56.8 | 75.7/54 | 77.5/59.1 | 80.8/59.7 | 74.7/51.1 | 77.4/58.7 |

Table 4: F1 and EM scores on the MLQA test set. XLTT with XLM-R denotes that XLTT adopts XLM-R as multilingual encoder to represent the input.

| F1/EM | English | Arabic | German | Greek | Spanish | Hindi | Russian | Thai | Turkish | Vietnamese | Mandarin | Avg |
|---|---|---|---|---|---|---|---|---|---|---|---|---|
| Multi-BERT (Hu, et al., 2020) | 83.5/72.2 | 61.5/45.1 | 70.6/54.0 | 62.6/44.9 | 75.5/56.9 | 59.2/46.0 | 71.3/53.3 | 42.7/33.5 | 55.4/40.1 | 69.5/49.6 | 58.0/48.3 | 58.7/49.4 |
| XLM (Hu, et al., 2020) | 74.2/62.1 | 61.4/44.7 | 66.0/49.7 | 57.5/39.1 | 68.2/49.8 | 56.6/40.3 | 65.3/48.2 | 35.4/24.5 | 57.9/41.2 | 65.8/47.6 | 49.7/39.7 | 59.8/44.3 |
| XLM-R (Hu, et al., 2020) | 86.5/75.7 | 68.6/49.0 | 80.4/63.4 | 79.8/61.7 | 82.0/63.9 | 76.7/59.7 | 80.1/64.3 | 74.2/62.8 | 75.9/59.3 | 79.1/59.0 | 59.3/50.0 | 76.6/60.8 |
| Translate-train Multi-BERT (Hu, et al., 2020) | 83.5/72.2 | 68.0/51.1 | 75.6/60.7 | 70.0/53.0 | 80.2/63.1 | 69.6/55.4 | 75.0/59.7 | 36.9/33.5 | 68.9/54.8 | 75.6/56.2 | 66.2/56.6 | 70.0/56.0 |
| Translate-train (multi-task) (Hu, et al., 2020) | 86.0/74.5 | 71.0/54.1 | 78.8/63.9 | 74.2/56.1 | 82.4/66.2 | 71.3/56.2 | 78.1/63.0 | 38.1/34.5 | 70.6/55.7 | 78.5/58.8 | 67.7/58.7 | 72.4/58.3 |
| Translate-test BERT-Large (Hu, et al., 2020) | 87.9/77.1 | 73.7/58.8 | 79.8/66.7 | 79.4/65.5 | 82.0/68.4 | 74.9/60.1 | 79.9/66.7 | 64.6/50.0 | 67.4/49.6 | 76.3/61.5 | 73.7/59.1 | 76.3/62.1 |
| XLTT with Multi-BERT (ours) | 85.2/73.9 | 63.7/46.6 | 72.3/55.7 | 64.2/46.5 | 77.1/58.4 | 61.0/47.7 | 72.9/55.1 | 44.9/35.8 | 57.7/42.4 | 71.1/51.3 | 60.2/50.4 | 66.4/51.3 |
| XLTT with XLM-R (ours) | 87.7/78.2 | 71.7/51.3 | 83.1/65.2 | 82.5/63.5 | 84.8/65.7 | 79.4/61.8 | 83.0/66.1 | 77.1/64.5 | 78.9/61.1 | 81.7/61.3 | 62.1/69.8 | 79.3/64.4 |

Table 5: F1 and EM scores on the XQuAD test set.

**MLQA** (Lewis et al., 2019) is a multi-way parallel extractive multilingual QA evaluation benchmark dataset including 7 languages detailed in Table 1.

**SQuAD v1.1** (Rajpurkar et al., 2016) is built based on Wikipedia articles in English. 536 high-quality hand-annotated articles are sampled and annotators formulate questions based on each individual paragraph requiring that the answer must be highlighted from the given paragraph. The data statistics of SQuAD v1.1 is shown in Table 2.

**DRCD** (Shao et al., 2018) **and CMRC2018** (Cui et al., 2019b) are two Chinese extractive reading comprehension datasets. The statistics are shown in Table 3.

### 5.3 Baselines

We turn to briefly introduce baseline models on MLQA and XQuAD to prove effectiveness of our approach, the size of models detailed in Table 6.

| Models | Model Size (million) |
|---|---|
| BERT-Large | 340 |
| Multilingual BERT | 110 |
| XLM | 270 |
| XLM-R | 550 |

Table 6: Size of Models

**Translate-Train:** This baseline model includes Translate-Train + Multi-BERT and Translate-Train + XLM. The training datasets such as SQuAD v1.1 are first translated into test languages to train Multi-BERT and XLM, and then evaluate them on test data.

**Translate-Test:** This baseline model represents Translate-Test + BERT-Large. Firstly, BERT-Large is trained using the SQuAD v1.1 English training data and the test data is translated into English. Then evaluate BERT_Large on the translated test set and translate the predicted answer into test languages for final evaluation.

**Multi-BERT:** Multi-BERT is a transformer-based language model pretrained on the Wikipedia of 104 languages using masked language modelling.

**XLM:** XLM uses a similar pretraining objective as multilingual BERT with a larger model, a larger shared vocabulary, and trained on the same Wikipedia data covering 100 languages, but XLM introduces additional translation language model.

**XLM-R:** XLM-R is similar to XLM but is trained on more than a magnitude more data from the web.

### 5.4 Evaluation

As far as we know, most of extractive RC tasks adopt EM and F1 metrics as evaluation measures. If prediction answer exactly matches the golden answer span, EM is equal to 1, otherwise 0. As for F1, a harmonic mean of precision and recall, each of which is calculated over instances within a language. The EM and F1 scores for each instance are averaged within each target language to obtain final F1 and EM scores, respectively. We take advantage of the multilingual evaluation script for the MLQA and XQuAD tasks to perform answer preprocessing operations such as stripping Unicode punctuation, standalone article stripping.

### 5.5 Results and Analysis

Table 4 and Table 5 show the results on the MLQA and XQuAD test sets, respectively. Overall, our XLTT with XLM-R as input multilingual encoder has the best generalization performance comparing the six baselines. To be specific, when comparing with the previous best model, our model achieves an average improvement of 5.1 in F1 and 4.5 in EM on the MLQA test set, and obtains an average improvement of 2.7 in F1 and 3.6 in EM on the XQuAD test set. The experimental results on the two multilingual ERC demonstrate effectiveness of our proposed multilingual ERC approach.

According to our preliminary simple analysis, we consider the improvement on MLQA and XQuAD owing to our proposed multilingual adaptive attention approach, data augmentation of constructed multilingual parallel corpora and multiple existing training datasets.

### 5.6 Ablation Study

To further determine the specific source of the contribution. We perform two ablation studies as follows. Table 7 shows the effect of training datasets on the improvement. We find that all three training datasets can improve the performance with different degrees using the new training framework.

| Pretrained XLTT (Avg F1/EM) | MLQA | XQuAD |
|---|---|---|
| XLTT with XLM-R | 77.4/58.7 | 79.3/64.4 |
| XLTT (w/o CMRC) | -1.3/-1.0 | -0.9/-0.8 |
| XLTT (w/o DRCD) | -1.1/-0.9 | -0.8/-0.8 |
| XLTT (w/o SQuAD) | -2.4/-1.8 | -2.1/-1.4 |
| XLTT (w/o CMRC, DRCD) | -1.5/-1.2 | -1.4/-1.1 |
| XLTT (w/o CMRC, SQuAD) | -2.9/-2.3 | -2.4/-1.7 |
| XLTT (w/o SQuAD, DRCD) | -2.6/-2.1 | -2.2/-1.6 |

Table 7: Ablation study for considering the effect of training datasets on XLTT model

To confirm the contribution of the multilingual parallel corpora and multilingual adaptive attention approach, we use SQuAD v1.1, CMRC2018 and DRCD to train XLM-R in cascading way and evaluate XLM-R on MLQA and XQuAD. The results in Table 8 show that multilingual parallel corpora can also improve the performance, but XLM-R is still behind our XLTT, which prove multilingual adaptive attention can also improve the performance.

| Models (avg F1/EM) | MLQA | XQuAD |
|---|---|---|
| XLM-R (w/ SQuAD) | 72.3/54.2 | 76.6/60.8 |
| XLM-R (w/ SQuAD, CMRC2018, DRCD) | 73.9/55.4 | 77.9/61.7 |
| XLTT with XLM-R (w/ SQuAD, CMRC2018, DRCD) | 77.4/58.7 | 79.3/64.4 |

Table 8: Effect of multilingual parallel corpora for the improvement

## 6 Conclusion

In this paper we propose a multilingual extractive RC modeling approach named XLTT, by a cross-lingual transposition rethinking approach for low-resource extractive reading comprehension using multilingual adaptive attention to model existing ERC training datasets in a multilingual context. Experimental results demonstrate the effectiveness of our multilingual ERC modeling approach.

## Acknowledgments

We would like to thank all anonymous reviewers for their thorough reviewing and providing constructive comments to improve this paper. This work was supported by the Ministry of Science and Technology via grant 2017YFB1401903 and 2018YFB1005101. This work was also supported by Beijing MoreHealth Technology Group Co. Ltd.